# ROCK HUNTING WITH MARTIAN MACHINE VISION

David Noever and Samantha E. Miller Noever

PeopleTec, Inc., Huntsville, Alabama, USA
david.noever@peopletec.com

## ABSTRACT

The Mars Perseverance rover applies computer vision for navigation and hazard avoidance. The challenge to do onboard object recognition highlights the need for low-power, customized training, often including low-contrast backgrounds. We investigate deep learning methods for the classification and detection of Martian rocks. We report greater than 97% accuracy for binary classifications (rock vs. rover). We fine-tune a detector to render geo-located bounding boxes while counting rocks. For these models to run on microcontrollers, we shrink and quantize the neural networks' weights and demonstrate a low-power rock hunter with faster frame rates (1 frame per second) but lower accuracy (37%).

## 1. INTRODUCTION

The Mars Perseverance rover and its Ingenuity helicopter offer opportunities for the application of computer vision [1-8]. Engineering tasks include obstacle avoidance [8], path planning, and robotic navigation [4-5]. Scientific tasks include rock classification [3,6], object detection, and elevation mapping. As an extreme example of edge computing, processors with autonomous decision-making make possible rapid mission planning [9] and auto-navigation. Its field-programmable gate arrays (FPGA, Virtex-5) perform maneuver controls for entry, descent, and landing, but receive new firmware to perform mobility visual processing [9]. When processing stereo-imagery at 0.67 frames per second and large 20-megapixel full-color CMOS sensors [9], Perseverance possesses a more robust self-driving capability with hazard avoidance compared to its traditional cycles to drive a meter, then wait while taking and processing a new image.

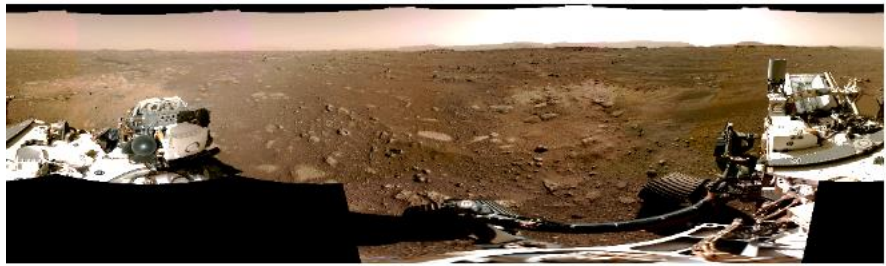

**Figure 1. NASA Perseverance Panorama on Sol 3. The large panorama consists of stitched together frames (142 total) taken using the Mastcam-Z stereo camera.**

Previous work applying machine learning to automate computer vision has featured site assessments from orbit [9-10]. From an overhead perspective, these trained algorithms act as rock counters. To overcome data latency and transmission delays, several authors [11-12] have explored deep learning and convolutional neural networks for object recognition. These rock counters [11] require training in complex planetary environments, where color contrast, edge detection, shadows, and textures challenge traditional computer vision applications. In the absence of machine learning, these conventional methodologies include algorithms [13] like Histogram Oriented Gradients (HOG), Scale-Invariant Feature Transform (SIFT), and Support Vector Machine (SVM). Typical accuracies for these approaches range from 60-75% [11] compared to 90+% classifier accuracy with deep learning. The original contributions of the present work include 1) dataset collection and object labeling with Perseverance's camera; 2) an object classifier built to distinguish rocks vs. rover (parts) with expert-level accuracy (>95%), and 3) a rock detector capable of running state-of-the-art deep learning algorithms (scaled Yolov4) on small edge processors. A simplified concept of operations for modern Martian vision applications shares features with autonomous driving on Earth. If one imagines the main task for self-driving balance safety against autonomous decision-making, then the key tasks include navigation (lane-following, turning, etc.) and hazard avoidance (terrain estimation, rock, and hole recognition). An additional opportunity for planetary rovers not shared with their terrestrial counterparts feature autonomous exploration and geological discovery. We discuss options for robotic autonomy and rock identification. The novel helicopter (Ingenuity) mission further offers scouting alternatives for route selection and rock hunting up to 980 feet per flight [9].

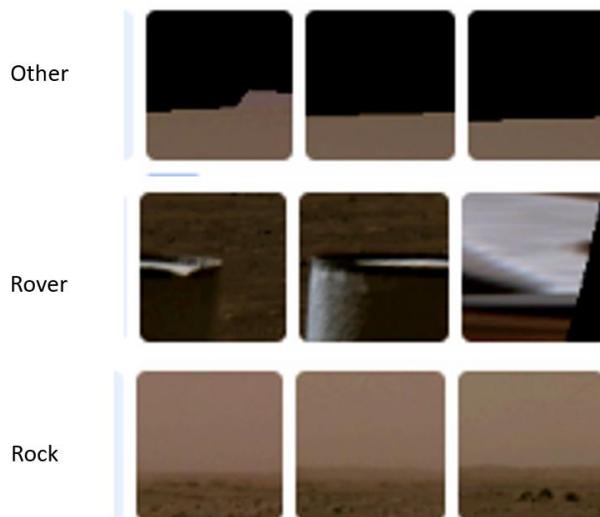

Figure 2 Example panels for multi-class identification

## 2. METHODS

NASA released their first post-processed panorama on 24 February 2021 [14] from their zoomable, stereo camera called Mastcam-Z [15]. On the third Martian day (Sol 3), the rover team published 142 individual images stitched together to form the 360-degree panorama (Figure 1). After downloading the panorama in PNG format (47 MB), we chipped the 75-megapixel image (16000x4721 pixels) into 1583 sub-images (224x224 pixels, Figure 2). To accommodate transfer learning approaches, we selected this chip size to speed up training for MobileNetV2 classifiers [16] and YOLOv4 scaled [17-18] methods. For all cases, we split the training, validation, and test sets into traditional 70:15:15 ratio of counts.

*Classifier.* We sorted the images into 3 categories ("other", "rock", and "rover") using human labelers and a python GUI program to separate them into separate folders [19]. The other category includes the Martian sky, soil without rocks, and distant mountains or craters. The rover category includes spacecraft parts as seen by the Mastcam-Z, including wheels and the rover body. A single rock in the image (or multiple) would categorize the chip as a positive case of the rock class. Using transfer learning and a partially pre-trained MobileNetV2 model, we first built a multi-class object recognition model using fine-tuning parameters (epochs=50, batch size =16, learning rate = 0.001). The premise behind transfer learning follows from the empirical observation that feature hierarchies built from large image datasets and different domains provide the core framework for analyzing new unrelated image classes. The benefits for classification include higher accuracy and a lower number of images per class.

*Detector.* Beyond classification *("Is there a rock or rover somewhere in the image?"),* we place bounding boxes and confidence intervals around each Martian rock. Of the original panorama, we chipped out 692 sub-images (416x416 pixels) that included rocks. We bounded each rock (1314) in the images using human labelers and a python GUI program [21] to annotate and format the pixel locations in rectangular boxes. Various authors [22] have offered a range of minimum image counts between 100 and 1000 per class, although some few-shot or zero-shot approaches [23] succeed for classification alone with fewer than 10 for uncommon objects. The scaled YoloV4 [17-18] provides a convenient starting point for fine-tuning a custom real-time object detector. In its balance of accuracy and speed, this model approaches the 15-30 frames per second typically regarded as video-quality for real-time in the literature. The model summary includes 235 layers, 52.5 million parameters, and 50.5 million gradients.

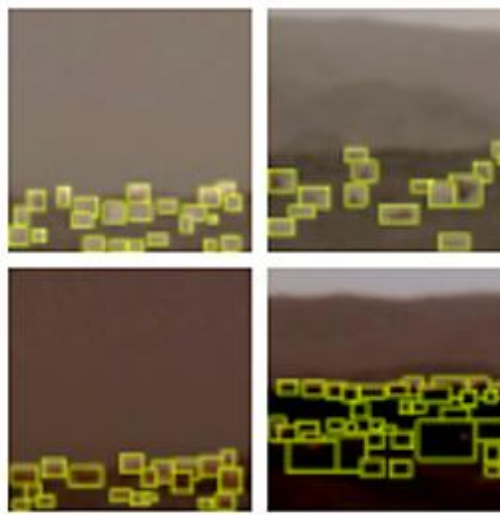

Figure 3. Bounding boxes around rocks in training data

## 3. RESULTS

Figure 4 shows the accuracy and confusion matrix for the basic classifier model. For typical human labelers, other studies have identified a classifier with accuracies about 95-97% as typical of expert-level over time, since without automation assistance human labelers tend to underperform over time for tedious tasks [24]. The size for this fine-tuned MobileNetV2 (14 MB) makes the model optimized to run at 30 frames per second on mobile phone processors. It's worth noting that given the risk acceptance for the Ingenuity helicopter, the Mars 2020 team accepted commercial phone-grade boards (Qualcomm Snapdragon 801). These processors exceed the rover's capabilities by several orders of magnitude but lack radiation hardening (and space legacy) for deep space travel [25].

|  | Rocks | Rover | Other |
|---|---|---|---|
| Rocks | 99.0% | 1.0% | 0.0% |
| Rover | 5.5% | 94.5% | 0.0% |
| Other | 0.0% | 1.3% | 98.8% |

**Figure 4** Confusion matrix for classifier

| On-device Performance | Inference Time (ms) | Peak RAM (K) | ROM Usage (M) | Accuracy |
|---|---|---|---|---|
| **Float 32 unoptimized** | 1718 | 957.2 | 1.60 | 97.0% |
| **Quantized (int8)** | 859 | 297.0 | 0.59 | 37.2% |

**Table 1**. Comparison of Image Classifiers Optimized for Microcontrollers

To explore the classifier performance, we ported the trained model to an Arduino Nano 33 BLE Sense device [26]. This low-cost device offers an Arm Cortex-M4 microcontroller running at 64 MHz with 1MB Flash memory and 256 KB of RAM. We deployed the model as an Arduino Sketch using the Edge Impulse platform [27]. Table 1 summarizes the on-device performance modeling for an unoptimized weight model (float32) and quantized (int8). The tradeoff between accuracy, speed, and size favors the float32 weights for accuracy to classify whether a test Martian image contains rocks or rover. Typical post-training gains from quantizing feature 75% smaller models with 200% faster execution. Given the color similarities, shadows, and lack of edge features, the accuracy for this example falls off sharply and would not represent a good deployment selection compared to the float32 version on microcontrollers.

Figure 5 shows a single frame of the rock detector (and counter) using the inference stage of YoloV4 on previously unseen test images. A summary video of rock detection is available online [28]. We constructed the video from multiple stacked images in a slideshow fashion with the bounding boxes placed around each identified rock. Using GPU (Nvidia Quadro M1200), this rock detector processes 40 frames per second (416x416 pixels) and counts the rock density ranging from 0-34 rocks per frame. The bounding box output optionally overlays both the object label and associated confidence intervals (0-1).

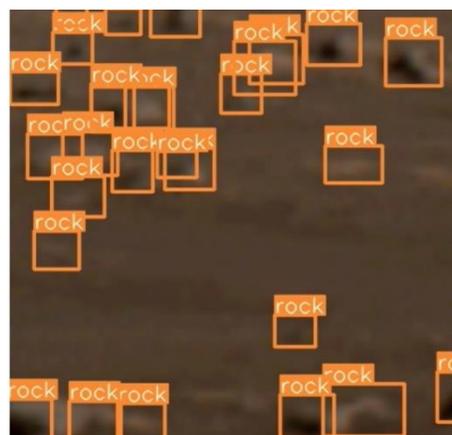

**Figure 5.** Bounding box detector for Mastcam-Z imagery

## 4. DISCUSSION AND CONCLUSIONS

This research explores the training of a classifier and detector for Martian rock discovery. We demonstrate greater than 97% accuracy for distinguishing rock versus rover from a single large image chipped into multiple frames and labeled for location (1314 total). We shrink the model with quantized weights and benchmark its performance for frame rate and accuracy on low-cost, low-power microcontrollers. Because of the low-contrast background of the initial panorama, the larger model (float32) outperforms the faster, smaller one. Future work can highlight the application of these trained models to automate the study of rock densities as the rover collects more imagery in different locations, lighting, and viewing angles.

## ACKNOWLEDGMENTS

The author would like to thank the PeopleTec Technical Fellows program for encouragement and project assistance.